%% file: transformer_mem.tex
\def\year{2022}\relax
\documentclass[letterpaper]{article} % DO NOT CHANGE THIS
\usepackage{aaai22}  % DO NOT CHANGE THIS
\usepackage{times}  % DO NOT CHANGE THIS
\usepackage{helvet}  % DO NOT CHANGE THIS
\usepackage{courier}  % DO NOT CHANGE THIS
\usepackage[hyphens]{url}  % DO NOT CHANGE THIS
\usepackage{graphicx} % DO NOT CHANGE THIS
\usepackage{natbib}  % DO NOT CHANGE THIS AND DO NOT ADD ANY OPTIONS TO IT
\usepackage{caption} % DO NOT CHANGE THIS AND DO NOT ADD ANY OPTIONS TO IT
\DeclareCaptionStyle{ruled}{labelfont=normalfont,labelsep=colon,strut=off} % DO NOT CHANGE THIS
\usepackage{algorithm}
\usepackage{algorithmic}

\usepackage{graphicx}
\usepackage{subfigure}

\usepackage{amssymb}
\usepackage{amsmath}
\usepackage{amsthm}
\usepackage{color}
\usepackage{xcolor}
\usepackage{comment}
\usepackage{multirow}
\usepackage{algorithm}
\usepackage{algorithmic}
\usepackage{multirow}
\usepackage[normalem]{ulem}
\usepackage{enumitem}
\setlist{nosep}
\setlist{noitemsep}
\usepackage{wrapfig}
\usepackage{lipsum}
\usepackage{fancyvrb}
\usepackage{bm}
\usepackage{array}
\usepackage{dsfont}

\newcommand{\method}{{\sc TMR}}

\newcommand{\electra}{{\sc ELECTRA}}

\newcommand{\inv}{\vspace*{-0.2cm}}
\newcommand{\sinv}{\vspace*{-0.1cm}}

\DeclareMathOperator{\sigmoid}{sigmoid}
\DeclareMathOperator{\softmax}{softmax}

\newcommand{\ph}[1]{\vspace{1mm} \noindent \textbf{#1.}}

\definecolor{battleshipgrey}{rgb}{0.52, 0.52, 0.51}

\usepackage{newfloat}
\usepackage{listings}
\floatstyle{ruled}
\newfloat{listing}{tb}{lst}{}
\floatname{listing}{Listing}
\title{Transformer with Memory Replay}
\author{
    Rui Liu, Barzan Mozafari
}
\affiliations{
    Computer Science and Engineering, University of Michigan, Ann Arbor\\
    \{ruixliu, mozafari\}@umich.edu
}

\usepackage{bibentry}
\begin{document}

\maketitle

\begin{abstract}
Transformers achieve state-of-the-art performance for natural language processing tasks by pre-training on large-scale text corpora. They are extremely compute-intensive and have very high sample complexity. Memory replay is a mechanism that remembers and reuses past examples by saving to and replaying from a memory buffer. It has been successfully used in reinforcement learning and GANs due to better sample efficiency. In this paper, we propose \emph{Transformer with Memory Replay} (TMR), which integrates memory replay with  transformer, making transformer more sample-efficient. Experiments on GLUE and SQuAD benchmark datasets show that Transformer with Memory Replay achieves at least 1\% point increase compared to the baseline transformer model when pretrained with the same number of examples. 
Further, by adopting a careful design that reduces the wall-clock time overhead of memory replay, we also empirically achieve a better runtime efficiency.
\end{abstract}

\input{introduction}

\input{related_work}

\input{technique}

\input{experiments}

\input{conclusion}
\section{Acknowledgments}
This work is in part supported by National Science Foundation and gifts from Google. The authors would like to thank anonymous reviewers for their insightful feedback and suggestions.

\bibstyle{aaai22}
\bibliography{bibliography}

%%%%%%%%%%%%%%%%%%%%%%%%%%%%%%%%%%%%%%%%%%%%%%%%%%%%%%%%%%%%

\end{document}

%% file: introduction.tex
\section{Introduction}

Transformers have achieved state-of-the-art performance on various natural language processing (NLP) tasks, such as sentimental anaylysis, paraphrase detection, machine reading comprehension, text summarization, question answering and so on~\cite{devlin2018bert,liu2019roberta,dai2019transformer,brown2020language}. 
The training of tranformers typically consists of two stages: pre-training and fine-tuning. 
Pre-training is the stage of training a generic model on an enormous corpus, such as Wikipedia to learn the representation inherent in understanding the natural language.
Fine-tuning is the stage of training a task-specific model that is initialized with pre-trained parameters on the dataset for the specific downstream task for just a few epochs. 
Each downstream task has a separate fine-tuned model, even though it is initialized with the same pre-trained parameters.
To get good generic representation, transformers are usually very large models with a huge number of parameters.
For example, OpenAI recently released GPT-3, which contains $175$ billion parameters~\cite{brown2020language}. 
Training these large-scale models is extremely compute-intensive and has very high sample complexity for the pre-training stage~\cite{devlin2018bert,clark2020electra}.
To make transformers more sample efficient, \citet{clark2020electra} proposed a new transformer model, called \electra{}, that consists of two modules: generator and discriminator. 
Both generator and discriminator are transformers, although generator is usually smaller in size. 
The discriminator is trained to predict whether each token in the corrupted input was replaced by the generator. 
They showed that their model substantially outperforms previous models, such as BERT and XLNet given the same amount of data. 

In this paper, we go one step further by integrating the memory replay mechanism into \electra{}, which we show can further improve the sample efficiency. 
Memory replay works by maintaining a fixed-size memory buffer that holds the most recent examples. 
It greatly improves the sample efficiency by enabling examples to be reused multiple times for training, rather than throwing away examples immediately after one-time usage.
By controlling the strategy on how examples are managed in the memory buffer (e.g., how to assign weights to examples), memory replay can be customized according to specific needs.
Memory replay mechanism has been successfully used in reinforcement learning~\footnote{The same concept is typically called \emph{experience replay} in reinforcement learning literature.} and GANs, 
because of its improvement on sample efficiency (i.e., requires less amount of samples to achieve the same accuracy)~\cite{schaul2015prioritized,fedus2020revisiting,wang2016sample,wu2018memory}.

\begin{figure}[ht!]
\centering
\includegraphics[width=8 cm]{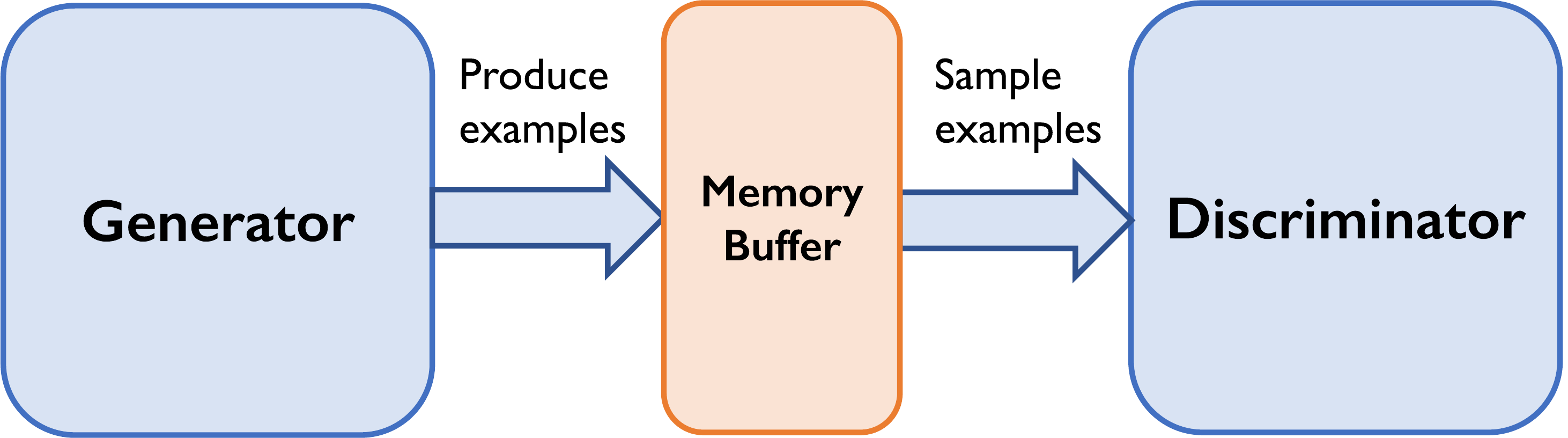}
\caption{The architecture diagram of our \emph{Transformer with Memory Replay} (\method{}). The generator produces corrupted examples, which are being saved to the fixed-size memory buffer. The discriminator samples examples from the memory buffer for training. High-quality examples are reused by the discriminator, thus making it more sample efficient to train the discriminator.}
\label{FIG:diagram}
%\vskip -0.2in
\end{figure}

Specifically, we keep the generator and discriminator from \electra{}, but add a memory buffer between them. 
In \electra{}, the generator produces corrupted text examples by trying to recover masked tokens, and discriminator takes the corrupted text examples as input and is trained to predict if each token is either original or replaced.
%Because generator and discriminator are being trained jointly, the distribution of corrupted examples produced by generator will shift, which can affect the training of discriminator.
Because the training objective of the generator is to mimic the original input sentences based on the masked ones, the generator will gradually drift away from its original purpose of providing random replacements to the input. 
Thus, the generator has the inevitable trend to produce lower and lower quality 
sentences as the generator is being optimized.
We use memory replay to alleviate the above issue, which we refer to as \emph{distribution drift} in this paper.
Our intuition is that, by saving examples produced by generator in the memory buffer, discriminator can reuse high-quality examples from the past, making it less sensitive to the distribution drift issue, and thus more sample efficient. 
We call this new transformer model \emph{Transformer with Memory Replay} (\method{}). 
Its architecture is illustrated in Figure~\ref{FIG:diagram}.
The detailed design of the memory replay would affect the performance of our new model. 
We will dicuss some design choices that lead to noticeable improvement on sample efficiency.
%It implies the importance of providing high-quality examples for training.
Going beyond the design choices discussed in this paper, we hope that our new model can be viewed as a general framework that provides an easy way to manipulate training examples via various memory replay designs, thus possibly achieving even better sample efficiency.

\begin{comment}
For the rest of our paper, we discuss existing works related to transformers and memory replay in Section ``Related Work". We elaborate on each component of our model and the distribution drift issue in Section ``Model Architecture", how to train them jointly in Section ``Joint Learning", and the strategies on weight update of the memory buffer in Section ``Memory Replay Weights". Experimental setup related to both pre-training and fine-tuning is discussed in Section ``Experimental Setup". Finally, we include the experimental results in Section ``Experimental Results".
\end{comment}

%% file: related_work.tex
\section{Related Work} \label{sec:related_work}
\ph{Transformer-based Models}
Before the transformer architecture came along, the dominant sequence models for NLP were based on complex recurrent neural networks, which are very hard to parallelize.
In~\cite{vaswani2017attention}, the \emph{transformer} architecture, solely based on attention mechanisms and easily parallelizable, was introduced to the NLP community, which has superior performance than traditional recurrrent neural networks. 
Ever since, transformer-based models have been the top performers in various NLP task competitions. 
For example, BERT~\cite{devlin2018bert} pre-trains a large transformer on unlabeled text corpora using the masked-language modeling task, achieving significant improvement on GLUE and SQuAD benchmark datasets.
This demonstrates the power of combining transformers and the self-supervised pre-training strategy. 
MASS~\cite{song2019mass} and UniLM~\cite{dong2019unified} extend BERT to the generation task by adding auto-regressive generative training objectives.
Instead of masking out input tokens as in a masked-language modeling task for pre-training, 
XLNet~\cite{yang2019xlnet} masks attention weights such that the input sequence is autoregressively generated in a random order.
\electra{} corrupts the input by replacing some tokens with plausible alternatives sampled from a generator, which they showed is more sample efficient.
It is also worth noting that
there is significant effort dedicated to improving the transformer architecture to achieve computation efficiency~\cite{kitaev2020reformer,kim2020fastformers,tay2020sparse,rae2019compressive,tay2020efficient} or to extend beyond the NLP domain~\cite{huang2018music,parmar2018image,karpov2019transformer,girdhar2019video,huang2020pop}.

\ph{Memory Replay}
Memory replay (or experience replay) is critical to deep reinforcement learning to achieve super-human performance~\cite{lin1992self,schaul2015prioritized,mnih2015human}.
It has been shown to improve sample efficiency and stability by storing and reusing past transitions~\cite{fedus2020revisiting}.
Some follow-up works have refined the basic memory replay mechanism~\cite{schaul2015prioritized} in various ways~\cite{horgan2018distributed,andrychowicz2017hindsight,sun2020attentive,luo2020dynamic,liu2019competitive}.
Several works have been trying to understand how memory replay works in the context of reinforcement learning. 
\citet{liu2018effects} study the effects of replay buffer size and mini-batch size on learning performance.
It has been reported that agent performance
is sensitive to the number of environment steps taken per
gradient step~\cite{fu2019diagnosing}. 
Sample efficiency can be improved by varying this ratio in combination with batch sizes~\cite{van2019use}.
In addition to reinforcement learning, \citet{wu2018memory} applies memory replay to GANs training in the task of learning new categories in a sequential fashion. They show that memory replay can prevent catastrophic forgetting~\cite{mccloskey1989catastrophic}, which is typically an issue in sequential learning.

%% file: technique.tex
\section{Transformer with Memory Replay}
\subsection{Model Architecture} \label{sec:model}
In this section, we describe the basic architecture of \emph{Transformer with Memory Replay} (\method{}). 
We keep the generator $G$ and discriminator $D$ from \electra{}, but add a memory buffer between them. 
%Next, we describe each one of them in detail.
We describe each of them and explain the distribution drift issue which is the reason for using a memory buffer as follows. 
%Details about each of them are described as follows.

\ph{Generator} 
The generator $G$ is a transformer network~\cite{vaswani2017attention} that maps a sequence of input tokens $\bm{x}^G=[x_1^G, \cdots, x_n^G]$
into a sequence of contextualized vector representations $h^G(\bm{x}^G)=[h_1^G, \cdots, h_n^G]$.
The input token sequence is obtained from the original text token sequence $\bm{x} = [x_1, \cdots, x_n]$ by masking out a random set of positions, i.e., replacing the original token with the \Verb+[MASK]+ token.
Typically, $15\%$ of input tokens are masked out randomly~\cite{devlin2018bert,clark2020electra}.
For any position $t$ where the corresponding input token $x_t^G$ is a \Verb+[MASK]+ token,
the generator outputs a probability for a particular token $x$ with a softmax layer given by
\begin{equation} \label{EQ:prob_G}
p_G(x|\bm{x}^G) = \frac{\exp\left( e(x)^T h^G_t  \right) }{\sum_{x'} \exp\left( e(x')^T h^G_t  \right) }
\end{equation}
where $e(x)$ is the embedding vector for token $x$.
The generator is trained with the masked language modeling (MLM) task, i.e., it learns to predict the original tokens for the masked out positions. 
Denote the set of masked out positions as $\bm{m} = [m_1, \cdots, m_r]$.
The loss function for the generator is
\begin{equation} \label{EQ:Loss_G}
L_G(\bm{x}^G, \theta_G) = \mathbb{E}\left(\sum_{i\in \bm{m}} -\log p_G(x_i^G|\bm{x}^G) \right).
\end{equation}

\ph{Discriminator}
Similar to the generator, the discriminator $G$ is also a transformer network mapping a sequence of input tokens $\bm{x}^M=[x_1^M, \cdots, x_n^M]$
into a sequence of contextualized vector representations $h^D(\bm{x}^M)=[h_1^D, \cdots, h_n^D]$.
Although the generator typically has the same model architecture as the discriminator, but smaller in size, 
the parameter values from the discriminator are used to initialize the task-specific model during the fine-tuning stage.
Existing work assumes the input to the discriminator comes directly from the output of the generator via the inference process, i.e., replace each \Verb+[MASK]+ token by a token that is sampled according to $p_G(\cdot|\bm{x}^G)$~\cite{clark2020electra}. 
Instead, we use corrupted examples $\bm{x}^M$ sampled from the memory buffer as the input to the discriminator, in order to mitigate the distribution drift issue which we will elaborate shortly.
For any position $t\in [1, n]$, the discriminator learns to predict whether the token $x^M_t$ is \emph{original} or \emph{replaced}.
A token is \emph{original} if it matches the token id from the original text input. 
Otherwise, a token is considered \emph{replaced}.
The probability that token $x_t^M$ is original is output by a sigmoid layer:
\begin{equation}
D(\bm{x}^M, t) = \sigmoid{}(w^T h_t^D)
\end{equation}
where $w$ is the learnable parameter to the sigmoid layer. 
The loss function of discriminator is 
\begin{equation} \label{EQ:Loss_D}
\begin{split}
L_D(\bm{x}^M, \theta_D) = &\mathbb{E}\left( \sum_{t=1}^n -\mathds{1}(x_t^M = x_t)\log D(\bm{x}^M, t) \right.\\
 &\left.-\mathds{1}(x_t^M \neq x_t)\log (1- D(\bm{x}^M, t))  \right).
\end{split}
\end{equation}

\ph{Distribution Drift}
We would like to point out the distribution drift issue of the generator, which is the key reason that motivates us to use the memory buffer.
Specifically, to minimize the negative maximum likelihood loss $L_G$ in Equation~\ref{EQ:Loss_G}, the generator will tend to mimic the original input sentences and generate examples that highly resemble the input, as the training is making more and more progress. 
This behavior will cause the generator to drift away from its original purpose of providing random replacements to the input. 
Imagine an extreme scenario where the generator is highly optimized based on Equation~\ref{EQ:Loss_G}, thus becoming perfectly capable of predicting the original token at each masked-out position.
This generator would produce the highly similar (if not exactly the same) sentence as the original one.
For example, if the original sentence $\bm{x}$ is ``The individual images in a film are called frames" with tokens ``image" and ``film" being masked out, the sentence produced by a highly optimized generator $\bm{x}'^M$  would probably be ``The individual images in a movie are called frames".
When this sentence $\bm{x}'^M$ is received by the discriminator, all the tokens are considered as \emph{original} except for the token ``movie", because ``movie" is the only token that is different from its corresponding one from the original sentence $\bm{x}$.
The discriminator is supposed to learn reasonable language semantics by solving a two-class optimization problem based on these labeled tokens. 
However, there are two issues with $\bm{x}'^M$ that hinder the discriminator's learning progress: 
\begin{enumerate}
\item the number of \emph{replaced} tokens is much less than the number of \Verb+[MASK]+ tokens (e.g., $50\%$ less here), resulting in insufficient number of \emph{replaced} tokens that would cause class imbalance problem for the discriminator; 
\item the token ``movie" that is considered as \emph{replaced} is essentially noisy for the discriminator, because the sentence $\bm{x}'^M$ itself is completely acceptable, and it would be more reasonable to consider ``movie" as \emph{original}.
\end{enumerate}
The above example shows how a low-quality sentence can hurt the discriminator's learning progress. 

On the other hand, an insufficiently optimized generator would probably produce an unacceptable sentence, e.g., ``The individual coma in a version are called frames". 
The tokens ``coma" and ``version" are considered as \emph{replaced} while other tokens are \emph{original}.
This is a better-quality sentence, because these labeled tokens imply that ``coma" and ``version" are not semantically related to other tokens from this sentence.
Unfortunately, the generator has the inevitable trend to produce lower and lower quality sentences as it is being optimized.
We call this phenomenon the \emph{distribution drift} of the generator. 
Memory replay is a mechanism that remembers and reuses past examples by saving to and replaying from a memory buffer.
It can be used to alleviate the distribution drift issue by replaying high-quality sentences from the memory buffer as input to the discriminator.
It is worth noting that memory replay can also be treated as an importance sampling technique~\cite{zhao2015stochastic,katharopoulos2018not}, if we directly use it to hold the training set. 
This would enable other transformer models such as BERT to make use of memory replay.
We do not consider this usage in this paper because memory replay is especially effective in handling dynamic example stream rather than static example set~\cite{lin1992self,schaul2015prioritized}.
Next we discuss the memory buffer that is inserted between the generator and the discriminator. 

\ph{Memory Buffer}
From $p_G$, we create a corrupted example $\bm{x}^M$ by replacing the \Verb+[MASK]+ token in $\bm{x}$ with a token randomly sampled according to $p_G(\cdot|\bm{x}^G)$ for all positions in $\bm{m}$. 
Any corrupted example $\bm{x}^M$ will be saved into the memory buffer.
We assign a real-valued weight to each example in the memory buffer, which indicates the importance of each example.
While the example importance is not directly accessible, we use different strategies to approximate it, which will be discussed in the next subsection.
The memory buffer has a fixed size $N$.
To be scalable when $N$ is large, we use a sum tree to maintain the example weights.
The memory buffer should support at least three operators: \Verb+add+, \Verb+update+ and \Verb+sample+.
The operator \Verb+add+ is used to add a new corrupted example $\bm{x}^M$ into the memory buffer.
The operator \Verb+update+ is used when we want to update the weight for an example already in the memory buffer.
The operator \Verb+sample+ is used to sample some examples from the memory buffer according to the weight distribution, i.e., the probability that any example is sampled is proportional to its weight.
Any of these three operators has $O(\log N)$ in complexity, because of the need to traverse the sum tree from the root to a leaf node. 
The operator \Verb+sample+ mentioned here is essentially a stochastic sampling method. 
In general, the probability of sampling example $i$ is 
\begin{equation}
P(i) = \frac{w_i^{\alpha}}{\sum_j w_j^{\alpha}}
\end{equation}
where $w_j$ is the weight for example $j$ and the hyperparameter $\alpha$ determines how much prioritization is used.
For example, $\alpha=0$ corresponds to uniform sampling, and $\alpha=\infty$ corresponds to greedy sampling (i.e., sample examples with the largest weights).
We illustrate more on how memory buffer works with some concrete examples in Figure~\ref{FIG:joint_learning_diagram}.

\subsection{Joint Learning} \label{sec:joint}
We learn the generator and discriminator jointly by minimizing the combined loss
\begin{equation}
\min_{\theta_G,\theta_D} \sum_{\bm{x}^G,\bm{x}^M} L_G(\bm{x}^G, \theta_G) + \lambda L_D(\bm{x}^M, \theta_D)
\end{equation}
where $\lambda$ is a scalar that balances the above two loss terms. 
We use the same optimizer, Adam with warmup, as in~\citet{clark2020electra} to iteratively minimize the combined loss.
Assume the mini-batch size is $K$.

\begin{figure}[t!]
\centering
\includegraphics[width=8 cm]{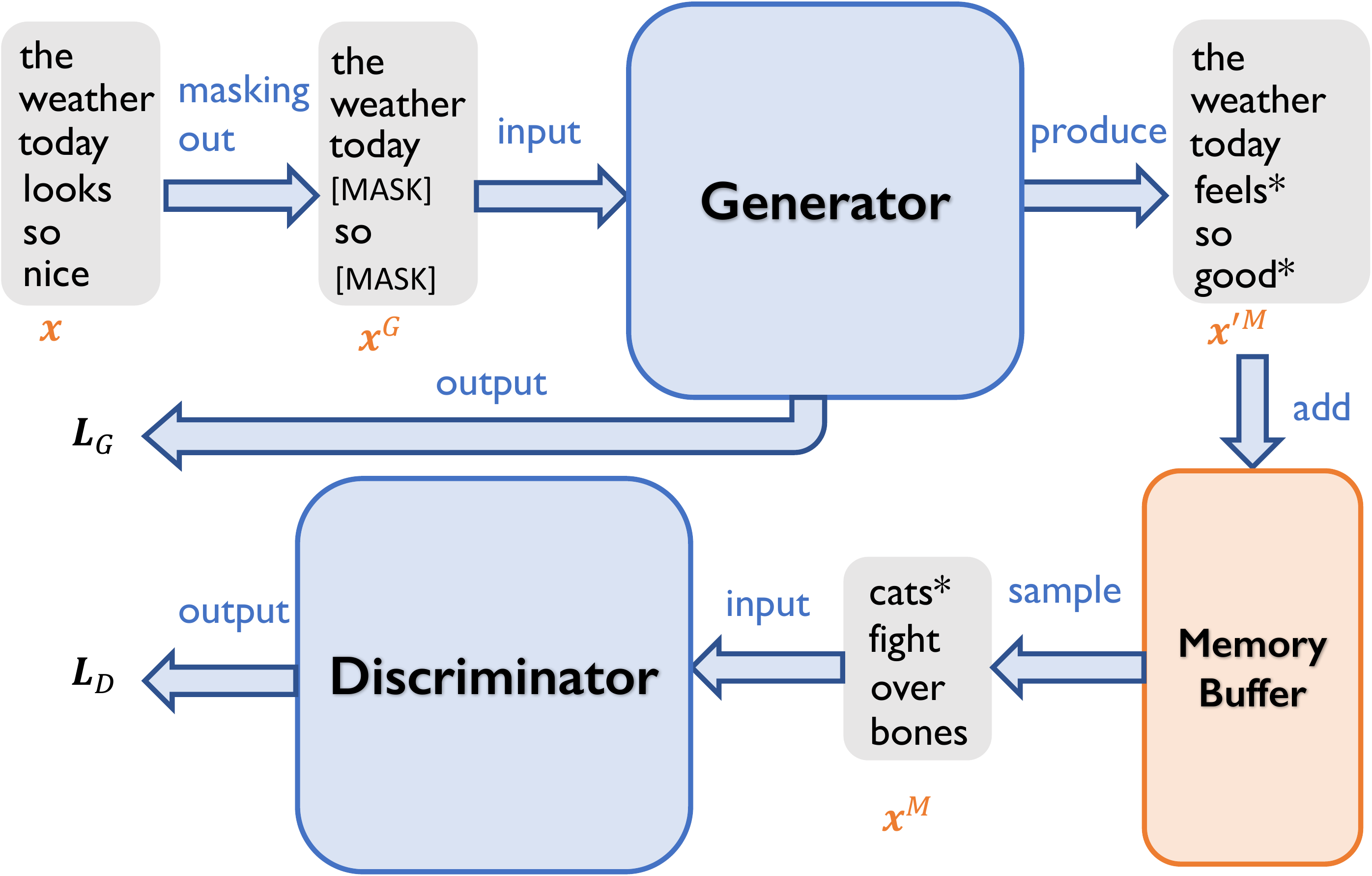}
\caption{Forward pass of \emph{Transformer with Memory Replay} (\method{}). For simplicity, mini-batch size is set to $1$ in the diagram to illustrate how an example is changed. We put $*$ right after a token to indicate this token has been replaced by the generator. Memory replay between the generator and discriminator helps the discriminator to learn more efficiently, by providing high-quality examples to it. The sentence ``the weather today feels* so good*" is saved to the memory buffer, and a better-quality sentence ``cats* fight over bones" which is sampled from the memory buffer is instead provided to the discriminator as input.}
\label{FIG:joint_learning_diagram}
\vskip -0.2in
\end{figure}

%As concrete examples to illustrate distribution drift, the generator produces the sentence ``the weather today feels so good", which greatly resembles the original sentence ``the weather today looks so nice", as an effect of minimizing its negative likelihood loss $L_G$. The produced sentence, however, is not very informative to the discriminator, because discriminator can not learn much from tokens ``looks" and ``nice" being replaced by ``feels" and ``good", respectively. Instead, the sampled sentence ``cats fight over bones" (``cats" might have replaced ``dogs") is more informative, because the discriminator will learn the important fact that ``cats" are not usually related to ``bones".}

\ph{Forward Pass}
Specifically, during the forward pass of each iteration, we sample a mini-batch of original text token sequence $\{\bm{x}_{(k)}\}_{k=1}^K$.
For each $\bm{x}_{(k)}$, a random set of positions is selected and the corresponding tokens are replaced by the \Verb+[MASK]+ token.
Thus, we get $\{\bm{x}^G_{(k)}\}_{k=1}^K$ by masking out random tokens from $\{\bm{x}_{(k)}\}_{k=1}^K$.
$\{\bm{x}^G_{(k)}\}_{k=1}^K$ are provided to the generator as input, 
and generator loss $L_G(\bm{x}^G, \theta_G)$ is computed according to Equation~\ref{EQ:Loss_G}.
In the meantime, from the generator output $p_G(x|\bm{x}^G)$, as in Equation~\ref{EQ:prob_G}, 
corrupted examples $\{\bm{x'}^M_{(k)}\}_{k=1}^K$ are created by replacing the \Verb+[MASK]+ token in $\bm{x}$ with a token randomly sampled according to $p_G(\cdot|\bm{x}^G)$.
The corrupted examples $\{\bm{x'}^M_{(k)}\}_{k=1}^K$ are saved into the memory buffer once created.
To get the input for the discriminator, we sample a mini-batch of examples $\{\bm{x}^M_{(k)}\}_{k=1}^K$ from the memory buffer using the \Verb+sample+ operator.
The detailed design of the memory buffer (e.g., how to assign/update example weights) will affect which examples get sampled.
Note that the sampled examples $\{\bm{x}^M_{(k)}\}_{k=1}^K$ are in general quite different from the examples $\{\bm{x'}^M_{(k)}\}_{k=1}^K$ that are just being saved to the memory buffer by the generator.
The discriminator then computes the loss $L_D(\bm{x}^M, \theta_D)$ according to Equation~\ref{EQ:Loss_D}.
The forward pass is illustrated in Figure~\ref{FIG:joint_learning_diagram} with mini-batch size as $1$.

\ph{Backward Pass}
During the backward pass, gradient with respect to generator (or discriminator) parameters is computed from the generator (or discriminator) loss. 
As with~\cite{clark2020electra}, we don't back-propagate the discriminator loss through the generator, which is difficult because of the sampling operation. 
Adam with warmup is used to update the generator (or discriminator) parameters based on the generator (or discriminator) gradient.

\subsection{Memory Replay Weights} \label{sec:weight}
Example weights play an important role in the memory replay mechanism.
At each iteration, the generator produces a mini-batch of new corrupted examples  $\{\bm{x'}^M_{(k)}\}_{k=1}^K$.
The operator \Verb+add+ is used to add these new examples into the memory buffer.
When a new example is being added to the memory buffer, an initial weight is assigned to it.
We then use the \Verb+sample+ operator to sample a mini-batch of examples $\{\bm{x}^M_{(k)}\}_{k=1}^K$ from the memory buffer and provide them to the discriminator as input.
The probability that an example is sampled is proportional to its weight.
After forward and backward pass, we use the \Verb+update+ operator to update the weights of the sampled examples based on feedback information from the discriminator.
We hope to increase the weights of high-quality examples, because they are more beneficial to training the discriminator.
In this way, these high-quality examples will be more likely to be sampled.

The strategy on how to assign and update example weights has great effect on which examples will be sampled as input to the discriminator, thus affecting its sample efficiency.
For each new example added to the memory, we assign its initial weight as the current average of the weights.
We also have tried other strategies to assign the initial weight, which are discussed in the experiments.
Since the memory buffer has a fixed size, it will eventually become full as we add more and more examples to it.
When we attempt to add a new example to a full memory, we evict the example with the lowest weight from the memory before adding the new example. 
The weight of each example indicates the importance of each example, which might change as the discriminator is being trained.
To keep the example weight as up-to-date as possible, we update the weight of each sampled example at the end of each iteration.
We discuss two different strategies on updating example weights.

\ph{Loss Difference}
We keep track of the discriminator loss $L_D$, as in Equation~\ref{EQ:Loss_D} for each example every time it is sampled. 
The loss difference of the current time and the previous time is used as the new weight. 
Note that, since computing the loss difference requires that an example gets sampled at least twice, we will not update the weight for any new example until the second time it is sampled.
In this case, we do not simply use the loss value as the new weight,
because the loss value typically decreases each time an example is sampled, 
so a smaller weight would not necessarily imply that it is less important.
Instead, it typically means that this example has been recently sampled. 
Our experiments also show that using loss value as the weight results in bad performance.

\ph{Gradient Norm}
Gradient norm has been shown to be the optimal weight for each example in importance sampling~\cite{zhao2015stochastic}. 
It demonstrates that per-example gradient norm can be a strong indicator for the example's importance.
Each time an example is sampled, we update its weight using the discriminator gradient norm for this example.
The drawback of using per-example gradient norm is that it incurs some overhead to compute the per-example gradient for each example in the mini-batch~\cite{goodfellow2015efficient}.
Recently, \citet{katharopoulos2018not} proposed an upper bound on gradient norm that can be computed efficiently. 
This upper bound has been shown to be a reasonable approximation on the gradient norm in training various neural network models~\cite{liu2020adam}. 
We also use this upper bound as the new weight.

%% file: experiments.tex
\section{Experimental Setup} \label{sec:setup}
\subsection{Pre-training}
We pre-train our model with two different sizes: a small model and a base model on English Wikipedia. 
The detailed hyperparameter values are included in the appendix. %listed in Table~\ref{TB:model_size}. 
As suggested by \citet{clark2020electra}, in addition to sharing the embedding table between input and output tokens for the generator, we also share the embedding table beween generator and discriminator. 
We use Adam with warmup to pre-train the models. 
The detailed setup is the same as \citet{clark2020electra} if not stated otherwise.
Specifically, we set $\epsilon=1e-6$, $\beta_1=0.9$ and $\beta_2=0.999$.
The mini-batch size is $128$ for the small model and $256$ for the base model. The memory buffer size $N$ is set to $1$k in our experiments.
%All of our experiments are conducted on a cluster of machines with NVIDIA P100 GPUs connected via $100$Gb/s InfiniBand Fabric.

\subsection{Fine-tuning}
We use two commonly used datasets as the benchmark to evaluate performance: 
General Language Understanding Evaluation (GLUE)~\cite{wang2018glue} and Stanford Question Answering Dataset (SQuAD)~\cite{rajpurkar2016squad}.

\ph{General Language Understanding Evaluation (GLUE)}
GLUE consists of eight tasks (i.e., MNLI, QQP, QNLI, SST, CoLA, STS, MRPC and RTE)\footnote{It is customary to exclude WNLI because it is difficult to beat even the majority classifier.}, each of which corresponds to a specific type of NLP problem. 
%Detailed information about the GLUE tasks is included in Table~\ref{TB:GLUE}. 
As with~\citet{clark2020electra}, our evaluation metrics are Spearman correlation for STS, Matthews Correlation for CoLA, and accuracy for other GLUE tasks. 
The average of these scores is reported.
Unless stated otherwise, results are on the dev set.
For fine-tuning, we only need the discriminator's parameters to initialize task-specific models~\cite{clark2020electra}.
Specifically, we use the final hidden vector $h_1^D\in \mathbb{R}^H$ of the discriminator corresponding to the first input token (\Verb+[CLS]+) as the aggregate representation~\cite{devlin2018bert}.
Note that $H$ is the hidden size. 
The only new parameters introduced during fine-tuning are classification layer weights $W\in \mathbb{R}^{C\times H}$, where $C$ is the number of labels. 
We use the standard cross-entropy loss for classification tasks:
$-\log\left( \softmax{}(h_1^DW^T)_{[class]} \right)$
where $[class]$ is the true class index.

\ph{Stanford Question Answering Dataset (SQuAD)}
The Stanford Question Answering Dataset (SQuAD v1.1) is a collection of $100$k crowd-sourced question/answer pairs~\cite{rajpurkar2016squad}.
Given a question and a passage from Wikipedia containing the answer, the task of SQuAD is to predict the answer text span in the passage.
As with the GLUE benchmark, for fine-tuning, we only need the discriminator's parameters to initialize task-specific models~\cite{clark2020electra}.
We introduce a start vector $S\in\mathbb{R}^H$ and an end vector $E\in \mathbb{R}^H$ to the task-specific model.
The probability of location $i$ being the start of the answer span is computed as a dot product between $h_i^D$ and $S$, followed by a softmax over all locations in the paragraph:
\begin{equation}
P_i^S = \frac{e^{S\cdot h_i^D}}{\sum_j e^{S\cdot h_j^D} }.
\end{equation}
Similarly, the probability of location $i$ being the end of the answer span is computed as a dot product between $h_i^D$ and $E$, followed by a softmax over all locations in the paragraph:
\begin{equation}
P_i^E = \frac{e^{E\cdot h_i^D}}{\sum_j e^{E\cdot h_j^D} }.
\end{equation}
The loss for fine-tuning is 
$-\log P_{[start]}^S - \log P_{[end]}^E$,
where $[start]$ and $[end]$ are the correct start and end positions.
The score of a candidate span from position $i$ to position $j$ is defined as $S\cdot h_i^D + E\cdot h_j^D$. 
The span with highest score where $i\leq j$ is used as the prediction.

\begin{figure*}[!t]
\inv
\centering
\subfigure[Small model on GLUE]{
\includegraphics[width=5 cm]{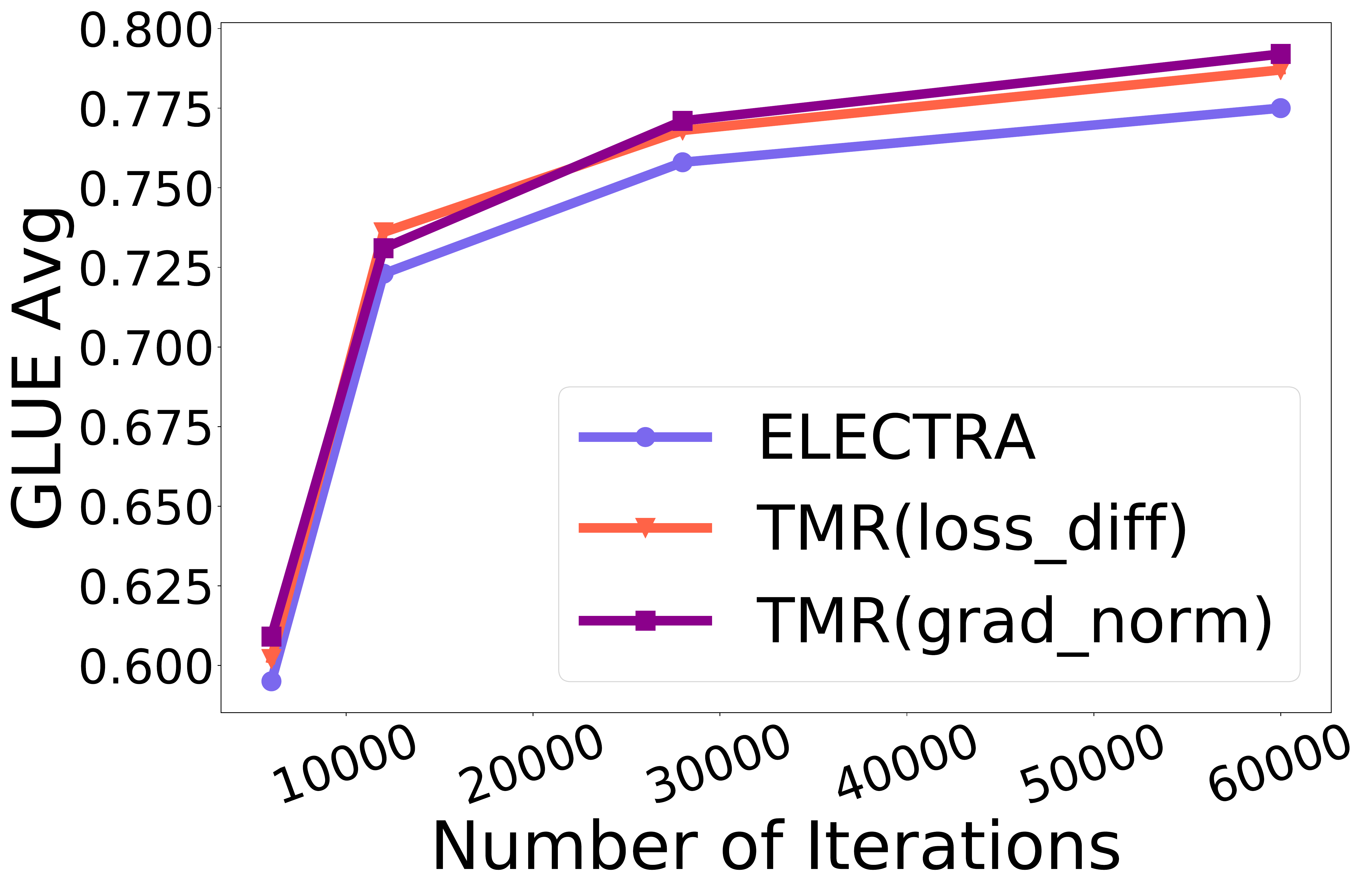}
}
\subfigure[Base model on GLUE]{
\includegraphics[width=5 cm]{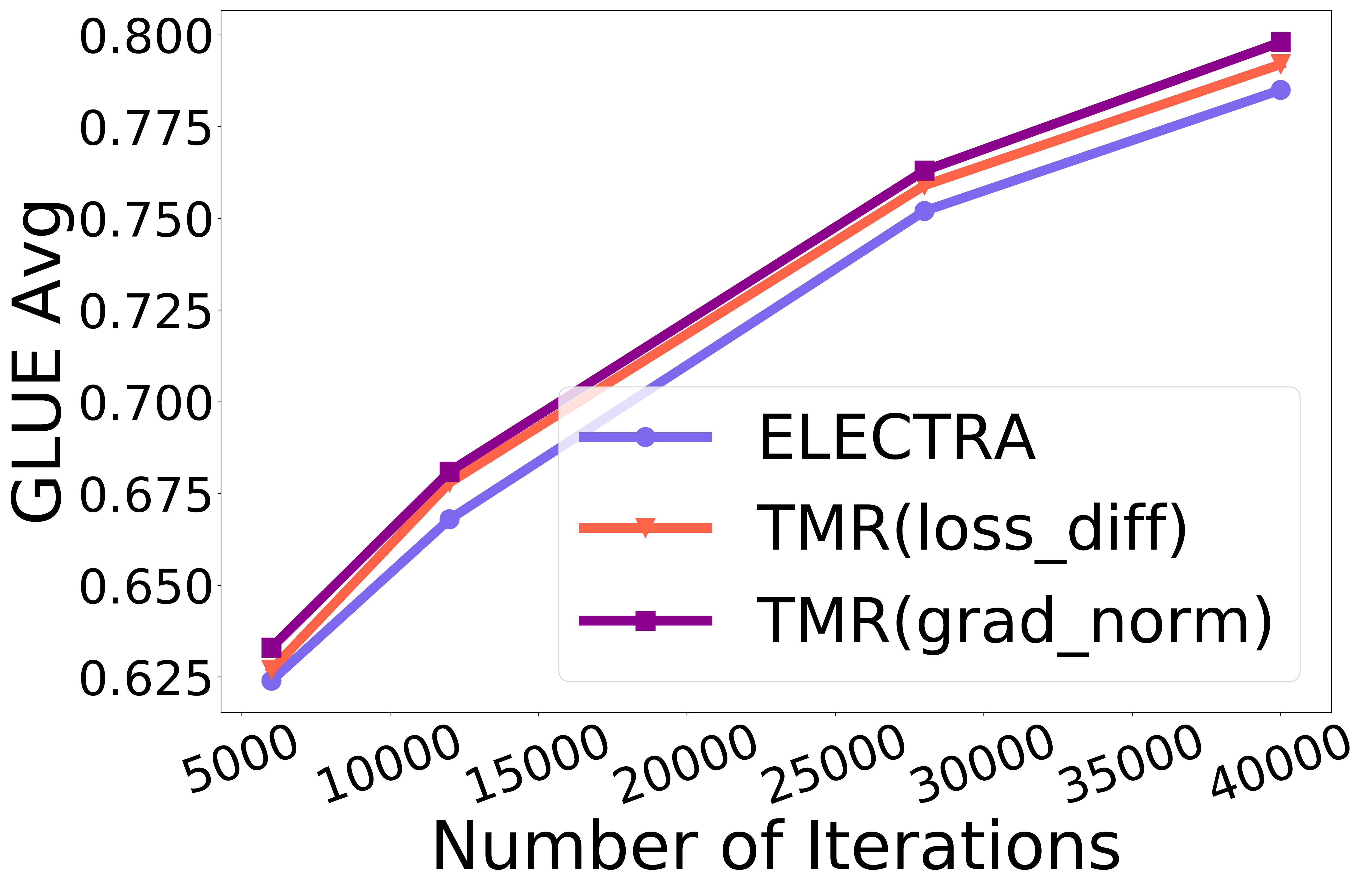}
}
\subfigure[Base model on SQuAD]{
\includegraphics[width=5 cm]{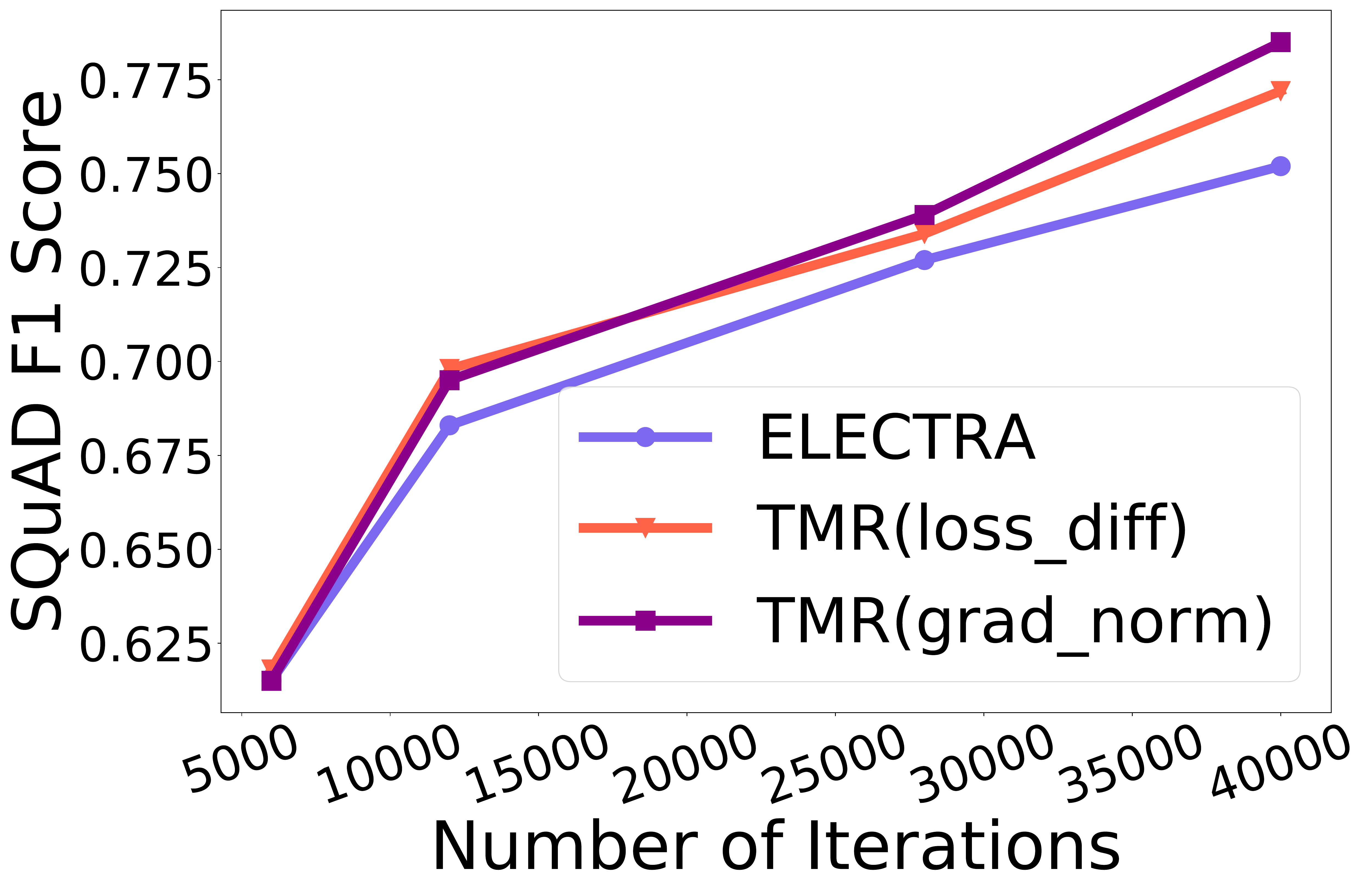}
}
\inv\sinv
\caption{Pre-training efficiency comparison. Our models (i.e., \method{}(loss\_diff) and \method{}(grad\_norm)) have better scores than baseline \electra{} when the pre-training has gone through enough iterations, because the memory replay needs time to accumulate enough high-quality examples.}
\inv
\label{fig:efficiency}
\end{figure*}

\begin{table*}[t!]
\small
\centering
%\vspace{0.2cm}
\scalebox{1}{
\begin{tabular}{|c|c|c|c|c|c|c|c|c|c|}
\hline
\textbf{Model} & \textbf{MNLI} & \textbf{QQP} & \textbf{QNLI} & \textbf{SST} & \textbf{CoLA} & \textbf{STS} & \textbf{MRPC} & \textbf{RTE} & \textbf{Avg.}\\
\hline
\hline
\electra{}  & $0.730$ & $0.881$ & $0.836$ & $0.862$ & $0.733$ & $0.832$ & $0.809$ & $0.584$ & $0.783$\\
\hline 
TMR(loss\_diff) & $0.753$ & $0.887$ & $0.846$ & $0.868$ & $0.737$ & $0.835$ & $0.811$ & $0.613$ & $0.794$\\
\hline
\end{tabular}
}
%\tofix{$C$ is the expected size of candidate set for \LSH{} and \PCA{}.} \barzan{where is C even used in this Table??}  
\caption{Results on GLUE of the small models after pre-trained for $200$k steps.}  
\label{TB:GLUE_test}
\end{table*}

\begin{comment}
\begin{table*}[t]
\small
\centering
\caption{Results on GLUE test set of the small models after pre-trained for $60$k steps.}
\vspace{0.2cm}
\scalebox{1}{
\begin{tabular}{|c|c|c|c|c|c|c|c|c|c|}
\hline
\textbf{Model} & \textbf{MNLI} & \textbf{QQP} & \textbf{QNLI} & \textbf{SST} & \textbf{CoLA} & \textbf{STS} & \textbf{MRPC} & \textbf{RTE} & \textbf{Avg.}\\
\hline
\hline
\electra{}  & $0.705$ & $0.881$ & $0.813$ & $0.846$ & $0.522$ & $0.739$ & $0.781$ & $0.561$ & $0.731$\\
\hline 
TMR(loss\_diff) & $0.713$ & $0.886$ & $0.817$ & $0.854$ & $0.540$ & $0.745$ & $0.773$ & $0.557$ & $0.736$\\
\hline
TMR(grad\_norm) & $0.724$ & $0.868$ & $0.823$ & $0.859$ & $0.545$ & $0.751$ & $0.788$ & $0.575$ & $0.742$ \\
\hline
\end{tabular}
}
%\tofix{$C$ is the expected size of candidate set for \LSH{} and \PCA{}.} \barzan{where is C even used in this Table??}    
\label{TB:GLUE_test}
\end{table*}
\end{comment}

\section{Experimental Results} \label{sec:results}
\subsection{Pre-training Efficiency}
To show that memory replay can improve the sample efficiency, we compare our models (i.e., \method{}(loss\_diff) and \method{}(grad\_norm)) with baseline \electra{} after being pre-trained with the same number of iterations (i.e., also the same number of examples, because we use the same mini-batch size for different methods). 
\method{}(loss\_diff) and \method{}(grad\_norm) are our model \emph{\underline{T}ransformer with \underline{M}emory \underline{R}eplay} using loss difference and gradient norm to update example weight, respectively.
Specifically, we pre-train different models to the same number of iterations using the same setting. 
We then fine-tune the models on GLUE and SQuAD to compare their performance.
The results are illustrated in Figure~\ref{fig:efficiency}.
When the number of iterations is small, our models have almost the same performance on both GLUE and SQuAD benchmarks.
As the number of iterations increases, we can see an obvious gap between our models and baseline. 
This is because, as more and more examples are processed, our models utilize the memory replay to remember more and more high-quality examples,
thus boosting the discriminator's learning speed.
On the other hand, the baseline \electra{} uses whatever examples that are produced by the generator to train the discriminator. 
As the training of the generator continues, the distribution of examples produced will also change due to the distribution drift issue, affecting the discriminator's training. 
%The results in Figure~\ref{fig:efficiency} are on the dev set.
%To verify that a similar trend can be observed on the test set, 
We list the detailed results for each GLUE task after $200$k steps in Table~\ref{TB:GLUE_test}.
As expected, after the same number of pre-training steps, our models \method{} have better performance than baseline across most of the GLUE tasks.  
Note that in the original paper~\cite{clark2020electra}, the baseline \electra{} is trained with much more iterations, getting slightly better GLUE scores than reported here.
We do not train the models for too many iterations because the max number of iterations used in our experiments suffices to demonstrate the benefits of the proposed memory buffer mechanism (especially given the increasing concern on the effect of excessive energy consumption on the environment~\cite{strubell2019energy,schwartz2019green}).
%the GLUE scores of the baseline \electra{} reported in our paper are slightly worse than the scores reported in the original paper~\cite{}. This is because the models are trained with much more iterations in the original paper. 

We also observe that \method{}(grad\_norm) often has better results than \method{}(loss\_diff). 
Theoretical analysis has been provided to show that gradient norm is the optimal weight for each example in importance sampling~\cite{zhao2015stochastic}.
Echoing this theoretical point, our observation empirically demonstrates that gradient norm is also a good choice for example weight in memory replay. 
Loss difference can be a bad approximation to gradient norm because model parameters might have changed significantly between adjacent visits to the same example. 
However, loss difference is a relatively cheap way to measure the example importance, which is especially beneficial in terms of the runtime efficiency as discussed right below.

\begin{figure}[!t]
%\inv
\centering
\includegraphics[width=5.5 cm]{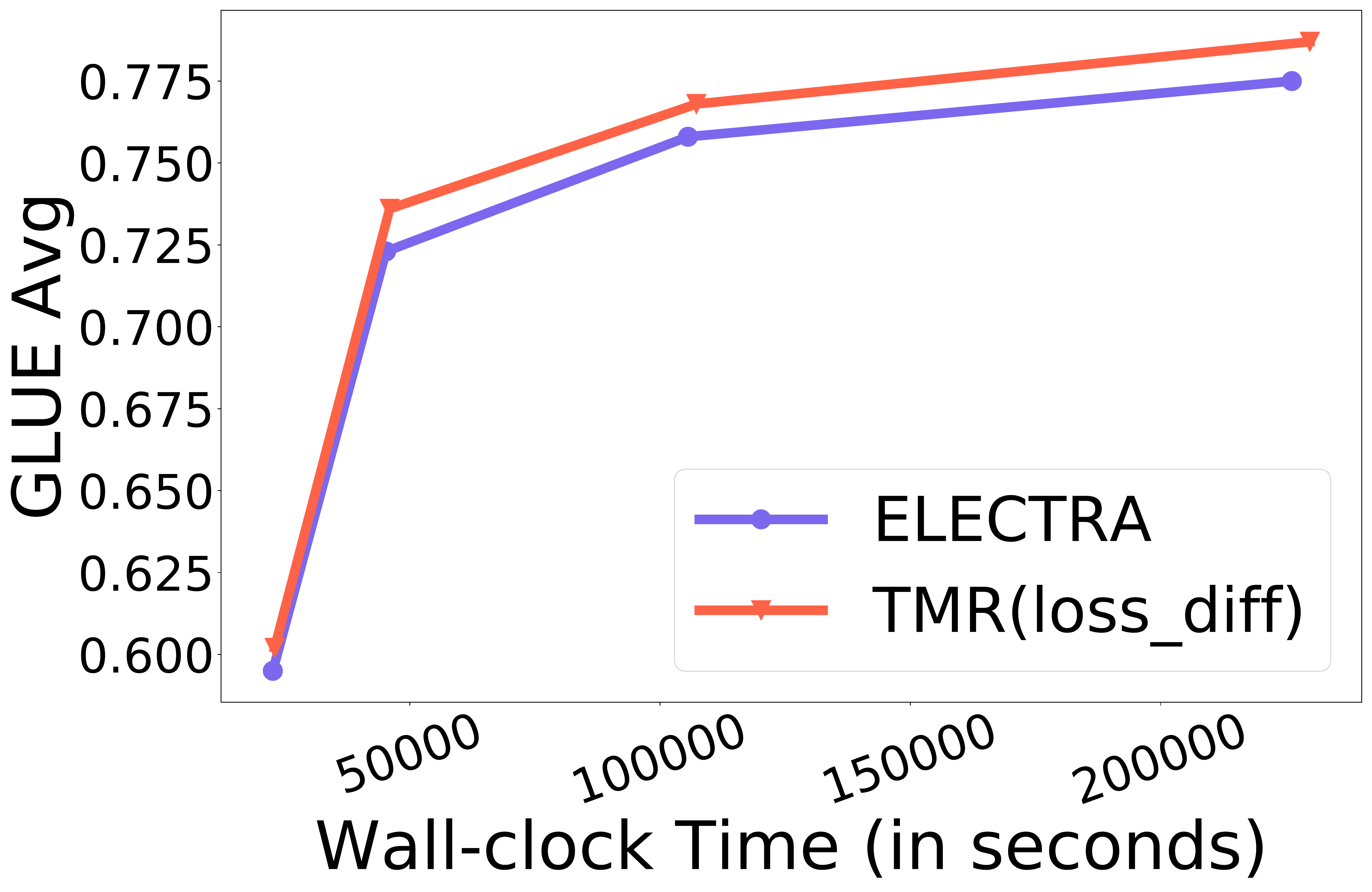}
\inv\sinv
\caption{Runtime efficiency comparison using small model on GLUE. \method{}(loss\_diff) has better runtime efficiency than the baseline \electra{}, because \method{}(loss\_diff) achieves better scores after pretaining for the same amount of wall-clock time.}
\inv
\label{fig:runtime_efficiency}
\end{figure}

\ph{Runtime Efficiency}
We also report the runtime efficiency in terms of the wall-clock time needed for training.
As shown in Figure~\ref{fig:runtime_efficiency}, using loss difference to measure the example importance reduces the wall-clock time overhead of the memory replay, leading to better runtime efficiency.

\subsection{Alternative Strategies for Weight Initialization}
We discuss some alternative strategies that we have tried but have worse performance.
We initialize the weight of a new example using the average of example weights in our previous experiments.
This strategy will assign the same weight to new examples that arrive at the same iteration (i.e., these new examples are in the same mini-batch).
The difference among examples is essentially ignored by this strategy.  
Intuitively, it might be better to assign weight to a new example using the weight of a similar example that is already present in the memory buffer.
Since each example is represented by a sequence of token indices $\bm{x}^M = [x_1^M, x_2^M, \cdots, x_n^M]$, which can be viewed as the feature vector of this example, the similarity of two examples can be measured by how similar their features are.
There are two alternative strategies that take into consideration the similarity among examples: \emph{Least Square Regression} and \emph{Linear Upper Confidence Bound}.
Before we dive into these two strategies, let us assume the memory buffer contains a set of examples and their corresponding weights, denoted as $\{\bm{x}^B_{(i)}, r_i \}_{i=1}^N$, where $r_i$ is the weight of the example $\bm{x}^B_{(i)}$ and $N$ is the capacity of the memory buffer.

\ph{Least Square Regression}
To determine the weights for a new example, we solve a \emph{Least Square Regression problem} (LSR) based on examples in the memory buffer and their weights: 
\begin{equation}
\theta^* = \mathop{arg} \mathop{min}_{\theta} \sum_{i=1}^N (\theta^T\bm{x}^B_{(i)} - r_i)^2.
\end{equation}
Then, we assign ${\theta^*}^T \bm{x}^M$ as the initial weight to the new example $\bm{x}^M$. 
Examples with different features will get different initial weights.

\ph{Linear Upper Confidence Bound}
Multi-armed bandit problem~\cite{bubeck2009pure,even2002pac} provides a general framework for studying efficient sampling methods. For example, it has been used to speed up optimization methods for model training~\cite{salehi2017coordinate,liu2020adam}, maximum inner product search~\cite{liu2019bandit}, hyperparameter search~\cite{li2017hyperband}, and model-related query execution~\cite{he2020method}. 
We cast the memory replay into a multi-armed bandit problem, and rely on a bandit method to initialize and update the example weights. 
Specifically, we treat each example as an arm.
In this way, sampling an example is equivalent to picking an arm to pull, 
and a high-quality example corresponds to an arm with high reward.
Because each example is represented as its feature vector, there are an infinite number of possible examples, implying an infinite number of arms.
To deal with an infinite number of arms, we resort to linear bandit, a special bandit setting where each arm is represented by a feature vector and reward is assumed to have a linear relationship with the arm feature vector.
We use a popular linear bandit method called \emph{Linear Upper Confidence Bound} (LinUCB), as described in Algorithm 1 of~\citet{chu2011contextual}. 
The Least Square Regression strategy will not change the inital weight of a new example until it gets sampled.
LinUCB can essentially keep refining initial weights of new examples as more and more examples are sampled.

\begin{figure}[!t]
%\inv
\centering
\includegraphics[width=5.5 cm]{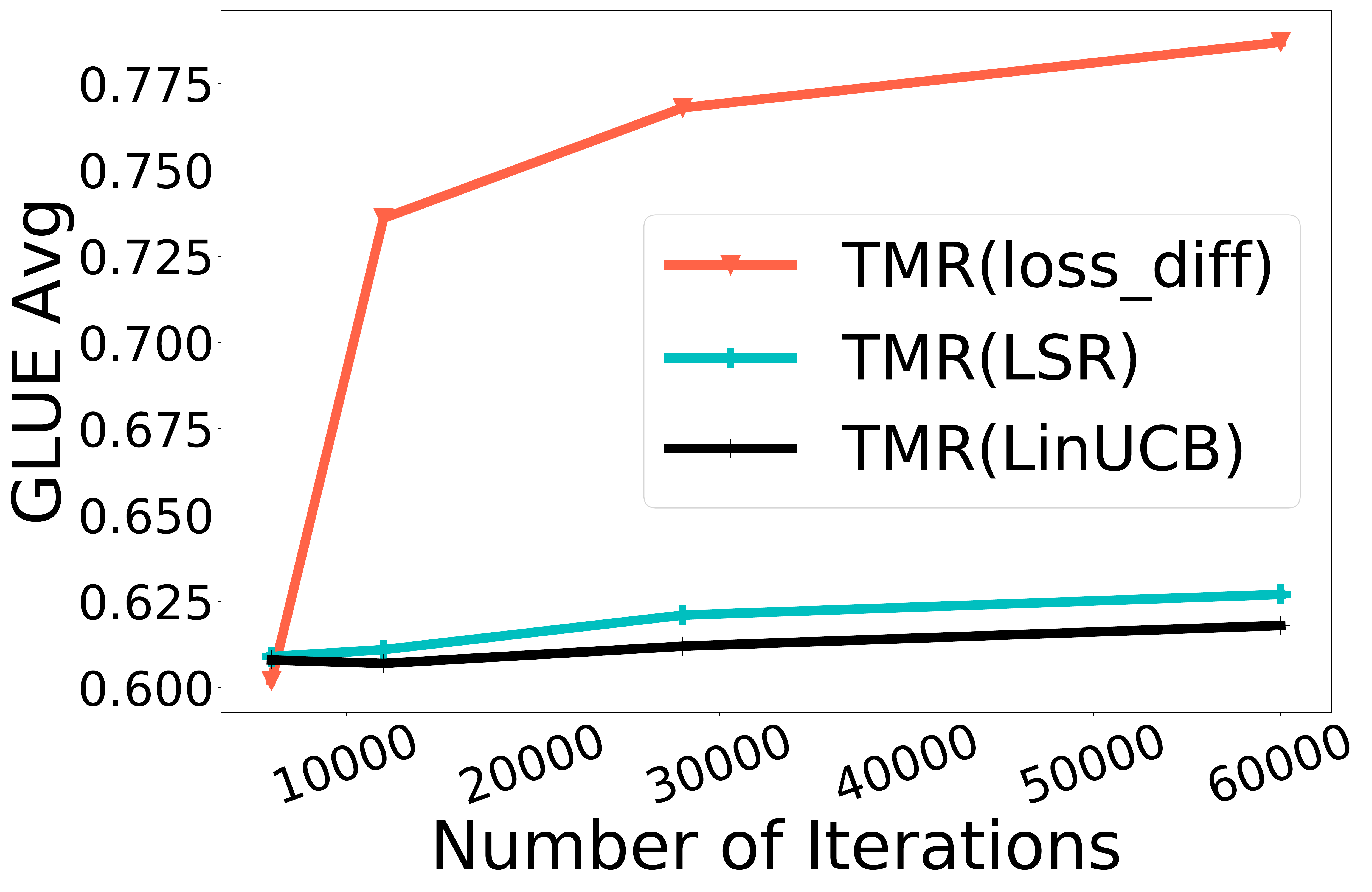}
\inv\sinv
\caption{Comparison of alternative strategies for weight initialization. TMR(LSR) and TMR(LinUCB) are much worse than TMR(loss\_diff) as the number of iterations increases.}
\inv
\label{fig:init_comparison}
\end{figure}

We did some experiments using LSR and LinUCB. The results are shown in Figure~\ref{fig:init_comparison}. 
We can see that both TMR(LSR) and TMR(LinUCB) are much worse than TMR(loss\_diff). 
These results are a bit counter-intuitive, because the idea of assigning weight to a new example using the weight of a similar example is reasonable.  
We suspect the reason why neither TMR(LSR) nor TMR(LinUCB) work is that the linear relationship between feature vector and weight/reward, which is assumed by both, does not hold in our memory buffer.
It would be a good future work direction to consider weight initialization strategies without linear relationship assumption.

\subsection{Runtime Cost}
We investigate the runtime cost for different models. In Table~\ref{TB:time_cost}, we list the wall clock time to finish $100$ iterations of training small models. 
The baseline \electra{} model has the lowest time cost, which is expected, because our models have the overhead of maintaining the memory buffer.
However, the time cost of TMR(loss\_diff) is only slightly larger than that of \electra{}.
This implies that the overhead of maintaining the memory buffer can be very small, because computing loss difference will not incur additional cost.
On the other hand, the time cost of TMR(grad\_norm) is way larger than the baseline. 
This is because accurately computing per-example gradient norm requires us to feed each example of a mini-batch separately to the model for both forward and backward pass.
This is usually very costly, because high parallelism provided by GPU cannot be fully utilized. 
TMR(grad\_bound) uses the gradient upper bound from~\citet{katharopoulos2018not} as an approximation to gradient norm.
It can significantly reduce the time cost of TMR(grad\_norm). 

\begin{table}[t]
\centering
%\vspace{0.2cm}
\scalebox{0.9}{
\begin{tabular}{|c|c||c|c|}
\hline
\textbf{Model} & \textbf{Time}&\textbf{Model} & \textbf{Time}  \\
\hline
\hline
\electra{}  & $377$ & TMR(grad\_norm) & $937$ \\
\hline 
TMR(loss\_diff) & $383$ & TMR(grad\_bound) & $432$ \\

\hline
\end{tabular}
}
%\tofix{$C$ is the expected size of candidate set for \LSH{} and \PCA{}.} \barzan{where is C even used in this Table??}   
\caption{Runtime cost comparison. The value in the table is the wall clock time (in seconds) to finish $100$ iterations.} 
\label{TB:time_cost}
\end{table}

%% file: conclusion.tex
\section{Conclusion}
We have proposed a new transformer model, called \emph{Transformer with Memory Replay} (\method{}), for improving the sample efficiency.
Our model integrates a memory replay mechanism into \electra{} by adding a memory buffer between the generator and discriminator.
Because the memory replay mechanism enables examples produced by the generator to be reused multiple times, 
the discriminator can be trained with high-quality examples, making it more sample efficient.
We also introduced two different strategies on how to update example weights (i.e., use loss difference or gradient norm). 
Our experiments have shown that our models achieve higher scores than baseline when pre-trained for the same number of iterations (i.e., using the same amount of examples).
We further demonstrated that the loss difference strategy leads to better runtime efficiency because of its low wall-clock time overhead.
%Not limited to the design choices we discussed in this paper, our new model can be viewed as a general framework where different designs of the memory replay mechanism is used to achieve even better sample efficiency.